\documentclass[10pt,twocolumn,letterpaper]{article}

\usepackage{cvpr}
\usepackage{times}
\usepackage{epsfig}
\usepackage{graphicx}
\usepackage{amsmath}
\usepackage{amssymb}
\usepackage{multirow,array}
\usepackage{algorithm}
\usepackage{algorithmicx}
\usepackage{algpseudocode}

\usepackage[pagebackref=true,breaklinks=true,letterpaper=true,colorlinks,bookmarks=false]{hyperref}
\cvprfinalcopy 

\def\cvprPaperID{2700} 

\ifcvprfinal\fi
\begin{document}
\bibliographystyle{plain}
\newcommand{\tabincell}[2]{\begin{tabular}{@{}#1@{}}#2\end{tabular}}
\title{End-to-end Video-level Representation Learning for Action Recognition}

\author{Jiagang Zhu$^{1,2}$, Wei Zou$^{1}$, Zheng Zhu$^{1,2}$\\
$^{1}$Institute of Automation, Chinese Academy of Sciences (CASIA)\\
$^{2}$University of Chinese Academy of Sciences (UCAS)\\
{\tt\small \{zhujiagang2015, wei.zou\}@ia.ac.cn, zhuzheng14@mails.ucas.ac.cn}
}


\maketitle

\begin{abstract}
From the frame/clip-level feature learning to the video-level representation building, deep learning methods in action recognition have developed rapidly in recent years. However, current methods suffer from the confusion caused by partial observation training, or without end-to-end learning, or restricted to single temporal scale modeling and so on. In this paper, we build upon two-stream ConvNets and propose Deep networks with Temporal Pyramid Pooling (DTPP), an end-to-end video-level representation learning approach, to address these problems. Specifically, at first, RGB images and optical flow stacks are sparsely sampled across the whole video. Then a temporal pyramid pooling layer is used to aggregate the frame-level features which consist of spatial and temporal cues. Lastly, the trained model has compact video-level representation with multiple temporal scales, which is both global and sequence-aware. Experimental results show that DTPP achieves the state-of-the-art performance on two challenging video action datasets: UCF101 and HMDB51, either by ImageNet pre-training or Kinetics pre-training. Codes are available at \url{https://github.com/zhujiagang/DTPP}
\end{abstract}


\section{Introduction}
In recent years, human action recognition has received increasing attention due to potential applications in human-robot interaction, behaviour analysis and surveillance. From traditional hand-crafted features based methods~\cite{MIFS15,Laptev05,IDT13,MOFAP16} to deep learning methods~\cite{TwoStream14,TSN16,TSN17,TLE16,DOVF17,DeepQuantization17,Gating17}, from small scale video datasets~\cite{hmdb51,ucf101} to large scale video datasets~\cite{Kinetics17,Karpathy14}, considerable progresses have been made in the community of human action recognition.

\begin{figure}[t]
\centering
\includegraphics[width=1\linewidth]{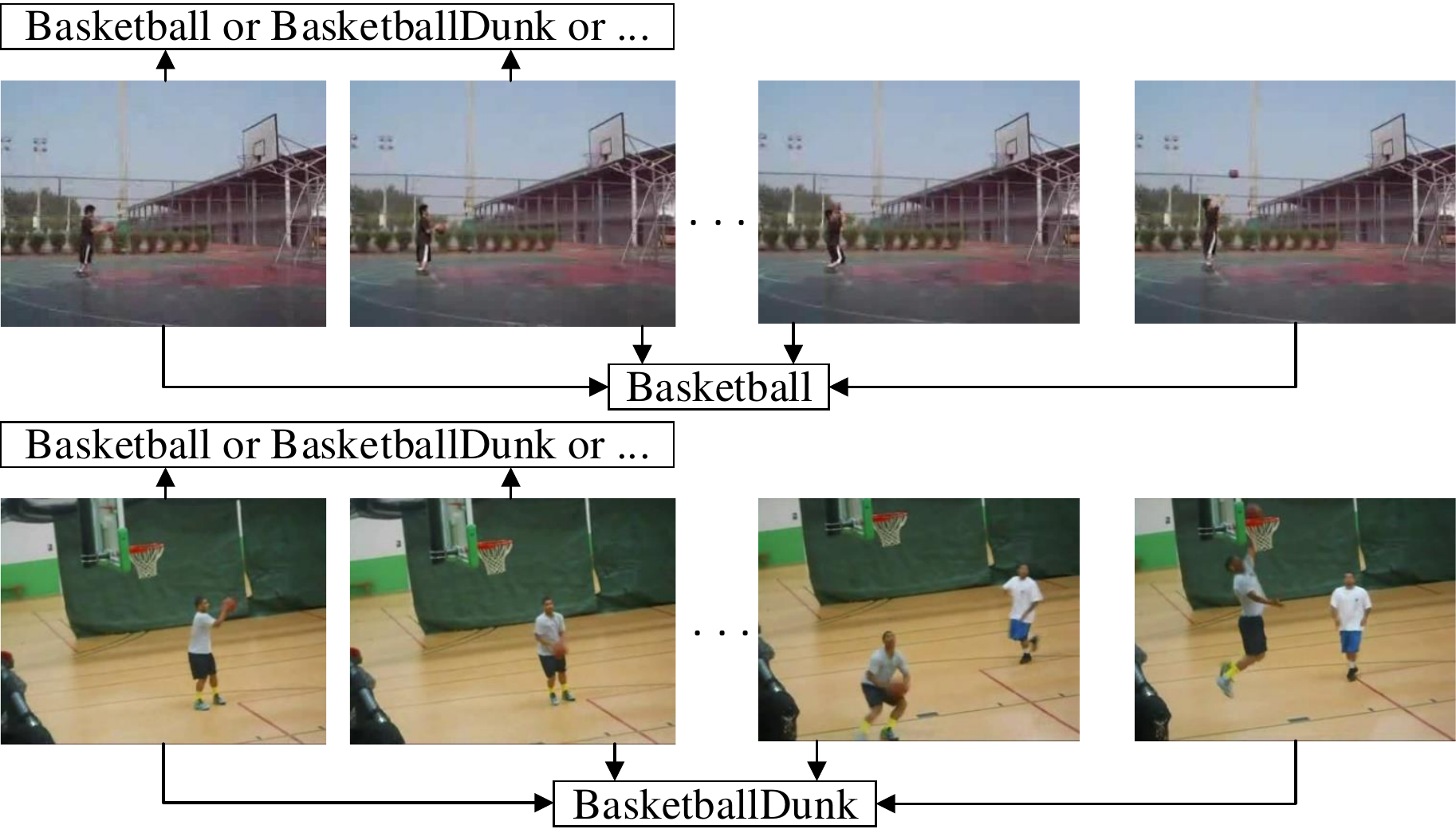}
\caption{The video frames of classes ``Basketball'' and ``BasketballDunk'' from UCF101. It is hard to tell the classes of partially observed videos until more frames are given.}
\label{Figure1}
\end{figure}

However, for the video representation building based on deep learning, there still remains some challenges to be solved and one of them is the long-range temporal structure modeling~\cite{TSN17,TLE16,DOVF17,DeepQuantization17}. As an image doesn't evolve over time like a video, it could be safely represented by its crop with the image-level label as the supervision~\cite{ImageNet12}. Given a video, early approaches tried to learn the frame-level feature~\cite{TwoStream14} or clip-level feature~\cite{C3D15}. At test time the frame-level predictions are averaged to get the video-level prediction~\cite{TwoStream14}. While the clip-level features are aggregated into a video representation for SVM classification~\cite{C3D15}. Recently, it is found that the classifier trained by single frame or short clip of a video\footnote {In this paper, we call it partial observation training} would confuse itself when it meets the look-alike frames/clips in the videos of different classes~\cite{ActionVLAD17,DOVF17}, as shown in Figure~\ref{Figure1}. To overcome the confusion caused by partial observation training, several works have been trying to learn the video-level representation~\cite{ActionVLAD17,convfusion16,LTC17,TLE16,DOVF17,DeepQuantization17,DynamicImage17}. They primarily follow two different pipelines: (1) aggregating the frame-level features of two-stream architectures into a fixed-size video-level representation~\cite{ActionVLAD17,TLE16,DOVF17,DeepQuantization17,DynamicImage17}, which employs complex encoding methods~\cite{bilinear2015,compactbilinear2016,FV10,VLAD10}; (2) applying 3D convolutions on long-term input frames to learn a spatio-temporal representation~\cite{convfusion16,LTC17}. However, their representations are either built with insufficient (\eg 3) frames/stacks sampled per video~\cite{TLE16}, or without end-to-end training~\cite{DOVF17,DeepQuantization17}, or restricted to single temporal scale modeling~\cite{DOVF17,DynamicImage17}, thus may not be discriminative enough and be sub-optimal. The representations of some methods are very high dimensional~\cite{ActionVLAD17,DeepQuantization17}, which make them hard to train and deploy. Moreover, compared to 2D convolutions, long-term 3D convolutions~\cite{LTC17} have to reduce the spatial resolution of input frames to relieve memory consumption, from which finer cues are easily lost.

To address these problems, a new video-level representation learning method is proposed in this paper, called Deep networks with Temporal Pyramid Pooling (DTPP) as illustrated in Figure~\ref{Figure2}. Our method aims to build the video-level representation in an end-to-end manner using enough frames sparsely sampled across a video. The trained model has the video-level representation and could make the video-level prediction both in training and testing instead of aggregating the frame-level predictions like some former two-stream style methods~\cite{TwoStream14,TSN16,TSN17,Gating17}. A temporal pyramid pooling layer is used to encode the frame-level features into the fixed-size video-level representation with multiple temporal scales, where a coarse-to-fine architecture is employed to capture the temporal structure of the human actions in videos.

Our method is built upon the two-stream style methods~\cite{TwoStream14,TSN16,TSN17,TLE16,Gating17}, where the spatial and temporal streams take as input RGB images and optical flow stacks respectively. The major difference is that our method aims to learn the video-level representation rather than the frame-level feature. By mapping between the video-level representation and prediction, it is expected to avoid the mistakes that the two-stream style models easily make due to partial observation training. Our method is also similar to some local encoding methods in action recognition~\cite{DOVF17,ORDER16,TPPC15}. The difference is that their methods are not trained end-to-end~\cite{DOVF17,ORDER16,TPPC15}, or restricted to single temporal scale learning~\cite{DOVF17}.

In summary, our contributions include: (a) We propose an end-to-end video-level representation learning method for action recognition, dubbed Deep networks with Temporal Pyramid Pooling (DTPP); (b) A temporal pyramid pooling layer is used to aggregate the frame-level features into the fixed-size video-level representation, which captures the temporal structure of videos in a multi-scale fashion; (c) DTPP achieves the state-of-the-art performance both on UCF101~\cite{ucf101} and HMDB51~\cite{hmdb51}, either by ImageNet~\cite{ImageNet_Pretraining_15} pre-training or Kinetics~\cite{Kinetics17} pre-training.

In the remainder of this paper, related works are discussed in Section~\ref{RelatedWorks}. Section~\ref{Methods} describes our proposed approach. In Section~\ref{Experiments}, our method is evaluated on the datasets. Finally, conclusions are given in Section~\ref{Conclusion}.

\section{Related works}
\label{RelatedWorks}

{\bf Hand-crafted video representation.} Before the surge of deep learning, the hand-crafted features are dominant in action recognition, such as Space Time Interest Points~\cite{Laptev05}, Cuboids~\cite{Cuboids}, Trajectories ~\cite{IDT13,DT13}. Among them, improved Dense Trajectory (iDT)~\cite{IDT13} is currently the basis of state-of-the-art handcrafted methods, which explicitly considers the motion features by pooling rich descriptors along dense trajectories and compensates camera motions. Then by encoding methods like BoW, Fisher vector, the descriptors are aggregated into the video-level representation. Some classification pipelines based on CNN are still built upon iDT, \eg trajectory-pooled deep-convolutional Descriptor (TDD)~\cite{TDD15}.

{\bf Frame-level feature by deep learning.} Simonyan \etal.~\cite{TwoStream14} designed two-stream ConvNets, which applies ImageNet pre-training for the spatial stream and optical flow for the input of the temporal stream. During training, it samples one RGB image or optical flow stack from each video. Temporal Segment Networks (TSN)~\cite{TSN16,TSN17} is an extension of the two-stream ConvNets by using deeper CNN and cross-modality pre-training. It aggregates several (3 or 7) frame-level predictions into a global video prediction during training and finally outperforms iDT~\cite{IDT13} by a large margin. C3D~\cite{C3D15} applies 3D convolutions on a short video clip with consecutive frames (typically 16). However, all these approaches build the frame/clip-level features instead of the video-level representations, thus suffer from the confusion caused by partial observation training. Motivated by the recent success of LSTM in sequence modeling~\cite{LSTM1,LSTM2,LSTM3,LSTM4,LSTM5}, there are also attempts that employ LSTM in video action recognition~\cite{UnsupervisedLSTM15,BeyondShort15,Attention16,TS-LSTM17}. Ng \etal.~\cite{BeyondShort15} trained LSTM with the CNN feature as input and aggregated predictions of all the time steps by linear weighting into the video-level prediction. LSTM has also been used to learn the sequence features of videos in unsupervised settings~\cite{UnsupervisedLSTM15} and to learn the attention model~\cite{Attention16}. However, due to its additional parameters as well as the differences between video frame and speech, text~\cite{TS-LSTM17}, LSTM has not shown its capacity in action recognition yet.

{\bf Video-level representation by deep learning.} Researchers have found the false label assignment problem~\cite{TSN16,TSN17,TLE16,DOVF17} caused by training from single frame or short clips of a video. Based on deep learning, several video-level representation learning methods have emerged recently. Varol \etal.~\cite{LTC17} explored the long-term temporal convolution to learn the video-level representation. However, because of the large amount of parameters of 3D convolution~\cite{LTC17,C3D15}, they had to downsample video frames to reduce memory consumption, thus be sub-optimal. Diba \etal.~\cite{TLE16} proposed Deep Temporal Linear Encoding (TLE), which achieves high performance by bilinear encoding~\cite{compactbilinear2016,bilinear2015}. However, the video-level representation of TLE contains insufficient information of a given video because it only samples 3 frames or flow stacks during training. Our method samples more (25) frames/stacks during training. There are also extensions~\cite{ActionVLAD17,DeepQuantization17} that insert traditional encoding methods, like Fisher vector~\cite{FV10} and VLAD~\cite{VLAD10} into deep CNN. They are designed to aggregate the local spatio-temporal features of CNN across the whole video into several clusters. The representations of these methods are either very high dimensional, \eg 32,768-D for ActionVLAD~\cite{ActionVLAD17}, 131,072-D for Deep Quantization~\cite{DeepQuantization17}, or without end-to-end training, \eg Deep Quantization~\cite{DeepQuantization17}. While the representation dimension of our method is lower than their methods, \ie 7168-D and it is also trained end-to-end. Moreover, our method explicitly considers multiple temporal scale video representation building, which is lacked in the single temporal scale methods, \eg Deep lOcal Video Features (DOVF)~\cite{DOVF17} and Dynamic Image Networks~\cite{DynamicImage17}.

\section{End-to-end video-level representation learning}
\label{Methods}

\begin{figure*}[t]
\centering
\includegraphics[width=0.9\linewidth]{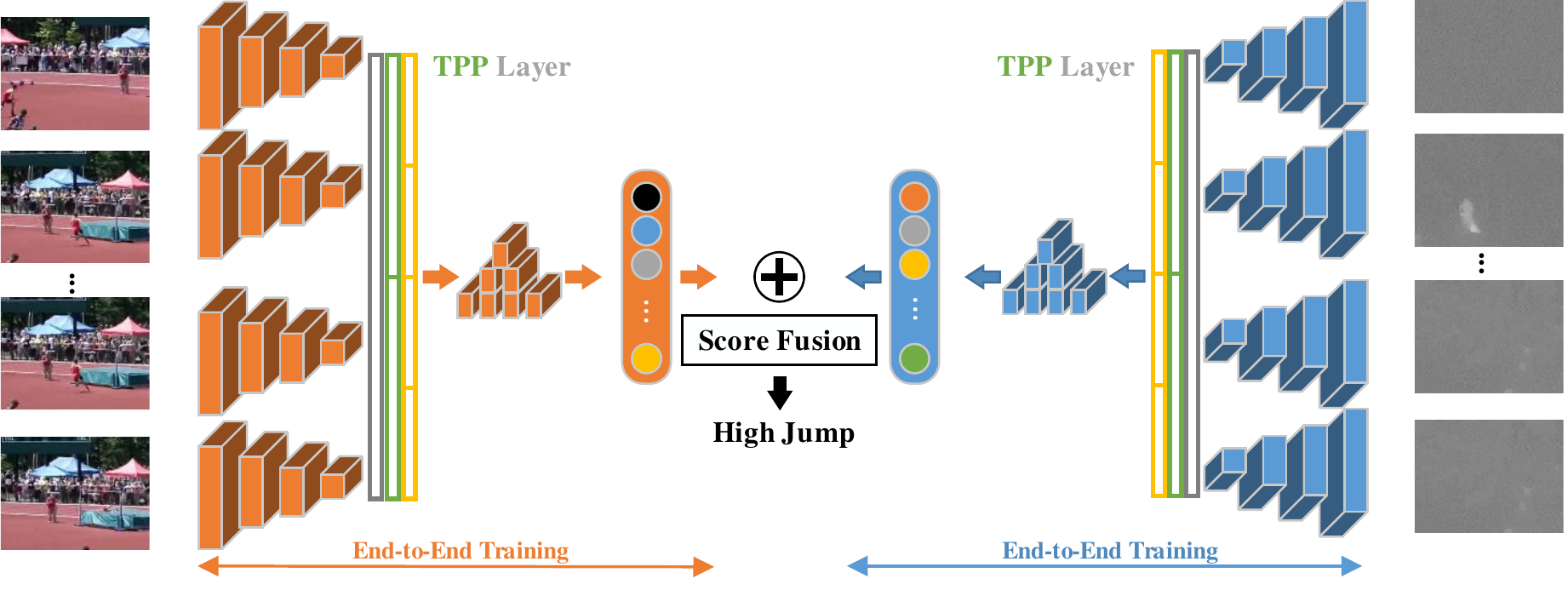}
\caption{Deep networks with Temporal Pyramid Pooling for video-level representation. The orange blocks denote the spatial stream and the blue blocks denote temporal stream. The spatial stream takes as input RGB images and the temporal stream takes as input optical flow stacks. The temporal pyramid pooling (TPP) layer for each stream aggregates the frame-level features into the video-level representation. Finally, the scores for the two streams are combined by weighted average fusion. The spatial ConvNets applied to different segments share weights and similarly for the temporal ConvNets. Each stream is trained end-to-end.}
\label{Figure2}
\end{figure*}

\begin{figure}[t]
\centering
\includegraphics[width=1\linewidth]{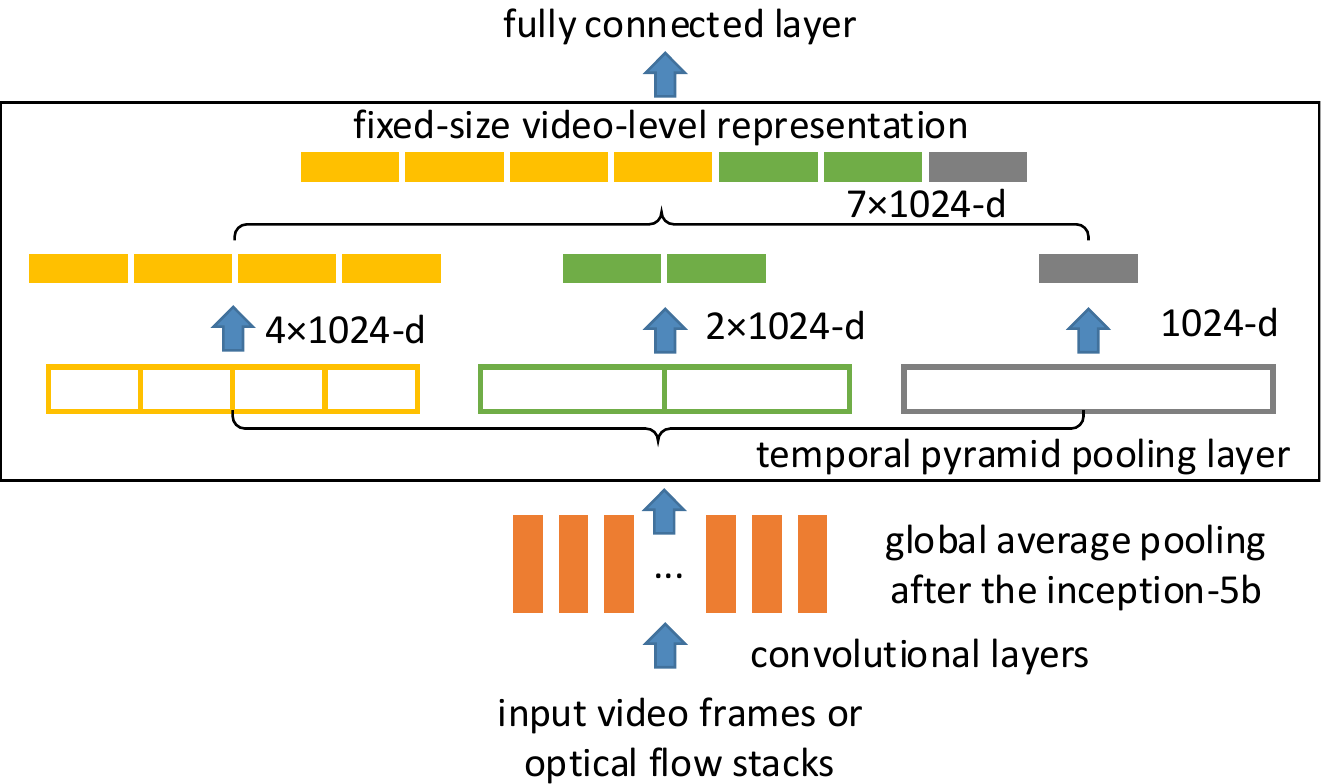}
\caption{An illustration of the Temporal Pyramid Pooling (TPP) layer. Here 1024 is the output dimension of the inception-5b, which is the last convolutional layer of BN-Inception. Note that the orange blocks denote the short-clip features. ConvNets on all short clips share parameters.}
\label{Figure3}
\end{figure}


\subsection{Overall architecture}
Figure~\ref{Figure2} shows the architecture of Deep networks with Temporal Pyramid Pooling (DTPP). Given a video $X$ consisting of $t$ frames, each frame is denoted as $x_i$ and the whole video as $X = \{x_1,x_2,\ldots,x_t\}$. Considering the redundancy of video frames and memory limitation, not all video frames are used to represent a video. Instead, each video is firstly divided into $T$ segments of equal durations and then a frame is sampled from each segment. These frames are fed to a ConvNet and the ConvNet computes the features for each frame independently. The obtained features are vectors~$\{S_1,S_2,\ldots,S_T\}$~where~$S_i\in{\mathbb{R}^d}, i=1,2,\ldots,T$, $d$ is the frame feature dimension.
To aggregate several frame-level features into a video-level representation, a temporal pyramid pooling (TPP) layer is placed on top of them. By end-to-end training, a fixed-size representation with multiple temporal scales is obtained.

\textbf{Network architecture.} The BN-Inception network~\cite{BN15} is applied as the backbone of DTPP. The BN-Inception has a depth of 33 layer, including 1 fully connected layer. As mentioned above, forward propagation is performed for each video frame independently. We extract the feature of the last convolutional layer after a global average pooling operation for each frame, which is a 1024-D vector. A temporal pyramid pooling layer is used to aggregate these frame-level features into a video-level representation, as shown in Figure~\ref{Figure3}. The number of pyramid levels of the temporal pooling layer is $K$ and the total number of bins is~$M={\sum}_{i=1}^{K}2^{i-1}$. In each temporal bin, the responses of each filter is pooled by a function $G$:~$\{S_i,S_{i+1},\ldots,S_j\}$${\to}$$Q_{i{\to}j}$. Different functions $G$ are investigated for the temporal pooling layer

\begin{itemize}
\item Average pooling:
\begin{equation}\label{eq1}
  Q_{i{\to}j}=(S_i{\oplus}S_{i+1}{\oplus}\ldots{\oplus}S_j)/(j-i+1)
\end{equation}
\item Max pooling:
\begin{equation}\label{eq2}
  Q_{i{\to}j}=max\{S_i,S_{i+1},\ldots,S_j\}
\end{equation}
\end{itemize}

Formally, the output of the temporal pyramid pooling layer $P$ is $M{\times}d$ dimensional, which is a video-level representation with $K$ pyramid levels
\begin{equation}\label{eq3}
\begin{split}
P_1&=[Q_{1{\to}T}]\\
P_2&=[Q_{1{\to}T/2},Q_{T/2+1{\to}T}]\\
   &\ldots\\
P_K&=[Q_{1{\to}T/(2^{K-1})},Q_{T/2^{K-1}+1{\to}T/(2^{K-2})},\\
    &\ldots,Q_{T{\times}(1-1/2^{K-1})+1{\to}T}]\\
  P&=[P_1,P_2,\ldots,P_K]\\
\end{split}
\end{equation}
where~$P{\in}\mathbb{R}^{M{\times}d}$~,~$P_i$~is the pyramid level $i$ video representation with dimension~$2^{i-1}{\times}d$.
Note that the coarsest pyramid level~$P_1$~covers the entire video. Each of the higher pyramid levels~$P_i,i=2,\ldots,K$~consists of the pooled bins aligned by the temporal order. Thus, the obtained representation is globally sequence-aware. In this way, DTPP is equipped with a degree of sequence regularization, which is lacked in some two-stream style methods, \eg TLE~\cite{TLE16}, TSN~\cite{TSN16,TSN17}, Deep Quantization~\cite{DeepQuantization17}. The video-level representation $P$ followed by a dropout layer is the input to the fully-connected layer, which outputs a video-level prediction given video $X$.

\textbf{Two-stream ConvNets.} Each input frame is rescaled by keeping the aspect ratio and its smaller side is resized into 256 pixels. RGB images and optical flow stacks are used as the inputs of two separate models respectively, \ie spatial stream and temporal stream. For spatial stream, each input frame size is 224$\times$224$\times$3. For temporal stream, each input frame size is 224$\times$224$\times$10, which is the stack of 5 consecutive optical flow fields~\cite{TSN16}.

\subsection{Training}
Consider a dataset of $N$ videos with $n$ categories $\{(X_i,y_i)\}_{i=1}^{N}$, where $y_i\in\{1,\ldots,n\}$ is the label. Our goal is to optimize the parameters of ConvNets to build the video-level representation $P$ by end-to-end training. Formally, the video-level prediction could be obtained directly

\begin{equation}\label{eq4}
  Y=\varphi(W_cP+b_c)
\end{equation}
where $\varphi$ is a softmax operation, ~$Y\in{\mathbb{R}^n}$. $W_c$ and $b_c$ are the parameters of the fully connected layer. In the training stage, combining with cross-entropy loss, the final loss function is

\begin{equation}\label{eq5}
  \mathcal{L}(W,b)=-\sum_{i=1}^{N}\log(Y(y_i))
\end{equation}
where $Y(y_i)$ is the value of $y_i$ th dimension of $Y$.
The back-propagation for the joint optimization of the $T$ segments can be derived as:

\begin{algorithm}[t]
\caption{Training}
\label{alg1}
\begin{algorithmic}
\\
\textbf{Input:}~$N$ videos with $n$ classes $\{(X_i,y_i)\}_{i=1}^{N}$, iteration number $Iter$\\
\textbf{Output:}~Parameters of ConvNets\\
Model initialization, $i = 0$.\\
\textbf{repeat}\\
~~~~~~1. Forward propagation and get $P$ with Eq.~(\ref{eq3}) \\
~~~~~~2. get video prediction $Y$ with Eq.~(\ref{eq4})\\
~~~~~~3. Back-propagation using Eqs.~(\ref{eq5}) and~(\ref{eq6})\\
~~~~~~4. $i = i + 1$\\
\textbf{until}~~$i = Iter$
\end{algorithmic}
\end{algorithm}

\begin{equation}\label{eq6}
  \frac{\partial{\mathcal{L}}}{dS_i}=\frac{\partial{\mathcal{L}}}{\partial{P}}\frac{\partial{P}}{\partial{S_i}}=\frac{\partial{\mathcal{L}}}{\partial{ P}}\sum_{j=1}^K\frac{\partial{P_j}}{\partial{S_i}}, i =1,2,\ldots,T\\
\end{equation}

\textbf{Model pre-training.} Two kinds of pre-training are used: ImageNet~\cite{ImageNet_Pretraining_15} pre-training and Kinetics~\cite{Kinetics17} pre-training. For the first one, the spatial stream is initialized by the model pre-trained on ImageNet and adapted by fine-tuning. To speed up convergence, we borrow the temporal stream of TSN~\cite{TSN16} trained on video action datasets to initialize our temporal model. For the second one, a large scale video dataset, Kinetics dataset~\cite{Kinetics17} is used to pretrain models. Kinetics dataset~\cite{Kinetics17} is built to explore the effect of transfer learning from large scale video dataset to small scale video dataset. The pre-trained TSN~\cite{TSN16,TSN17} models provided by Xiong~\cite{TSN_Kinetics17} are used, which are firstly adapted to our DTPP framework and further fine-tuned on small video action datasets~\cite{hmdb51,ucf101}. The convolutional parameters of two streams are transferred from the pre-trained models and the fully connected layers are randomly initialized. The learning procedure is summarized in Algorithm~\ref{alg1}.

\textbf{Training detail discussions.} $T=$~25 RGB images and optical flow stacks are sampled from each video by default. In later ablation studies, different number of frames/stacks are also sampled from each video to explore the effect of number of training frames.

\subsection{Inference}
For each video during testing, 25 RGB images and optical flow stacks are sampled. Meanwhile, the crops of 4 corner and 1 center, and their horizontal flippings are obtained from the sampled frames. From one of these 10 crops, 1 video-level prediction is obtained. Corresponding 10 video-level predictions are averaged to get the final prediction. All predictions are obtained without the Softmax normalization.

\textbf{Model fusion}
For the fusion of the predictions from two streams, weighted averaging scheme is used, where the fusion weight for each stream is 0.5 by default. The scores are fused without the Softmax normalization.

\section{Experiments}
~\label{Experiments}
In this section, the evaluation datasets and the implementation details of our approach are firstly introduced. Then, quantitative and qualitative results on these datasets are reported.
\subsection{Datasets and implementation details}
\textbf{Datasets.} Experiments are conducted on two challenging video action datasets: UCF101~\cite{ucf101} and HMDB51~\cite{hmdb51}. The UCF101 dataset contains 13,320 videos with 101 action classes. The HMDB51 dataset contains 6,766 videos with 51 action classes. For both datasets, standard 3 training/testing splits are used for evaluation. The mean average accuracy over three testing splits is finally reported for the comparison with other approaches.


\textbf{Implementation details.} The mini-batch stochastic gradient descent algorithm is used to learn the model parameters, where the batch size is set to 128\footnote{ Note that the batch size for updating gradients is different from the {\tt batch$\_$size} for each forward and backward pass in {\tt train.prototxt}. The {\tt batch$\_$size} in {\tt train.prototxt} for both streams is 4. For the spatial stream, {\tt iter$\_$size} in {\tt solver.prototxt} is 32, 1 GPU is used, so the batch size (for updating gradients) $=$ {\tt batch$\_$size} $\times$ {\tt iter$\_$size} $\times$ {\tt n$\_$gpu} = 128. For the temporal stream, {\tt iter$\_$size} in {\tt solver.prototxt} is 16, 2 GPUs are used, so the batch size (for updating gradients) $=$ {\tt batch$\_$size} $\times$ {\tt iter$\_$size} $\times$ {\tt n$\_$gpu} = 128.} videos. The L2 norm of gradients is clipped at 40 and momentum term is set to 0.9. Instead of decreasing the learning rate according to a fixed schedule, the learning rate is lowered by a factor of 10 after validation error saturates. Specifically, the learning rate for spatial stream is initialized as 0.01, ended at 0.00001, while the learning rate for temporal stream is initialized as 0.001, ended at 0.00001. The dropout ratio of the dropout layer for the video-level representation of both streams is set to 0.8. The same data augmentation techniques are used for all frames/stacks sampled from one video, including location jittering, horizontal flipping, corner cropping and scale jittering~\cite{TSN16}. The optical flows are computed using the method of~\cite{TVL1}. All experiments are implemented with Caffe~\cite{caffe14}. One NVIDIA TITAN X GPU is used for training the spatial stream and two for the temporal stream. OpenMPI\footnote{https://github.com/yjxiong/caffe} is used for the data parallel of the temporal stream.

\subsection{Ablation studies}
In this section, different aspects of learning the video-level representation with our approach are investigated by experiments. These experiments are all conducted on the split 1 of the UCF101 dataset.

\textbf{Temporal pyramid pooling.} The number of pyramid levels and kernel of temporal pooling are selected according to the performance of the network. Max pooling is used as the temporal pooling kernel by default. As shown in Table~\ref{Table2}, with 1 pyramid level, \ie global max pooling over 25 frame-level features, the two-stream accuracy of DTPP has achieved 94.3\%. With further finer scale pooling, the performance is enhanced and it finally plateaus at 4 pyramid levels. Note that the representation of 4 pyramid levels has nearly twice the dimension of that of 3 pyramid levels. A single pyramid level with 3 segments is also tested, whose accuracy is 0.3\% lower than that of 2 pyramid levels with the same dimension, which demonstrates the merits of multi-scale temporal pooling. The temporal pyramid pooling layer with average pooling kernel and 3 pyramid levels is also verified, which gains lower accuracy than that with max pooling kernel. Thus, the temporal pooling layer with 3 pyramid levels and max pooling kernel is selected both for the spatial and temporal streams by default.

\begin{table}
\begin{center}
\begin{tabular}{|c|c|c|c|}
\hline
    & Spatial & \tabincell{c}{Temporal} &  \tabincell{c}{Two-stream} \\
\hline\hline
1              & 88.2    & 88.2                    & 94.3 \\
1,2            & 89.6    & 88.9                    & 94.8 \\
1,2,4          & 89.7    & 89.1                    & 94.9 \\
1,2,4,8        & 89.7    & 88.9                    & 94.9 \\
3              & 89.3    & 88.6                    & 94.5 \\
1,2,4(Ave)     & 87.8    & 88.4                    & 94.4 \\
\hline
\end{tabular}
\end{center}
\caption{Exploration of different number of temporal pyramid level and pooling kernel for DTPP on UCF101 (split 1), accuracy({\%}).}
\label{Table2}
\end{table}

\textbf{Effect of end-to-end training.} The effect of end-to-end training on the video-level representation is explored. For direct comparison, TSN~\cite{TSN16} with BN-Inception, a frame-level feature learning method is chosen to do experiments. We also design a variant of TSN, ``TSN+TPP'', which is adapted from TSN by adding a TPP layer with 3 pyramid levels and max pooling kernel. This variant is only fine-tuned in the final fully connected layer, while the parameters of the convolutional layers are initialized by the trained reference models on UCF101 (split 1) and kept fixed. As shown in Table~\ref{Table3}, though without end-to-end training, TSN+TPP improves the accuracy of TSN by 0.9\%, showing the importance of building the video-level representation. Our method not only exceeds TSN, but also outperforms TSN+TPP by 0.5\%, demonstrating the merits of end-to-end training in building the video-level representation.

\begin{table}
\begin{center}
\begin{tabular}{|c|c|c|c|}
\hline
    & Spatial & \tabincell{c}{Temporal} &  \tabincell{c}{Two-stream} \\
\hline\hline
TSN~\cite{TSN16} & 85.7    & 87.9                    & 93.5 \\
TSN+TPP          & 88.0    & 88.9                    & 94.4 \\
DTPP             & 89.7    & 89.1                    & 94.9 \\
\hline
\end{tabular}
\end{center}
\caption{Exploration of end-to-end training for DTPP on UCF101 (split 1), accuracy({\%}).}
\label{Table3}
\end{table}

\begin{figure}[t]
\centering
\includegraphics[width=0.9\linewidth]{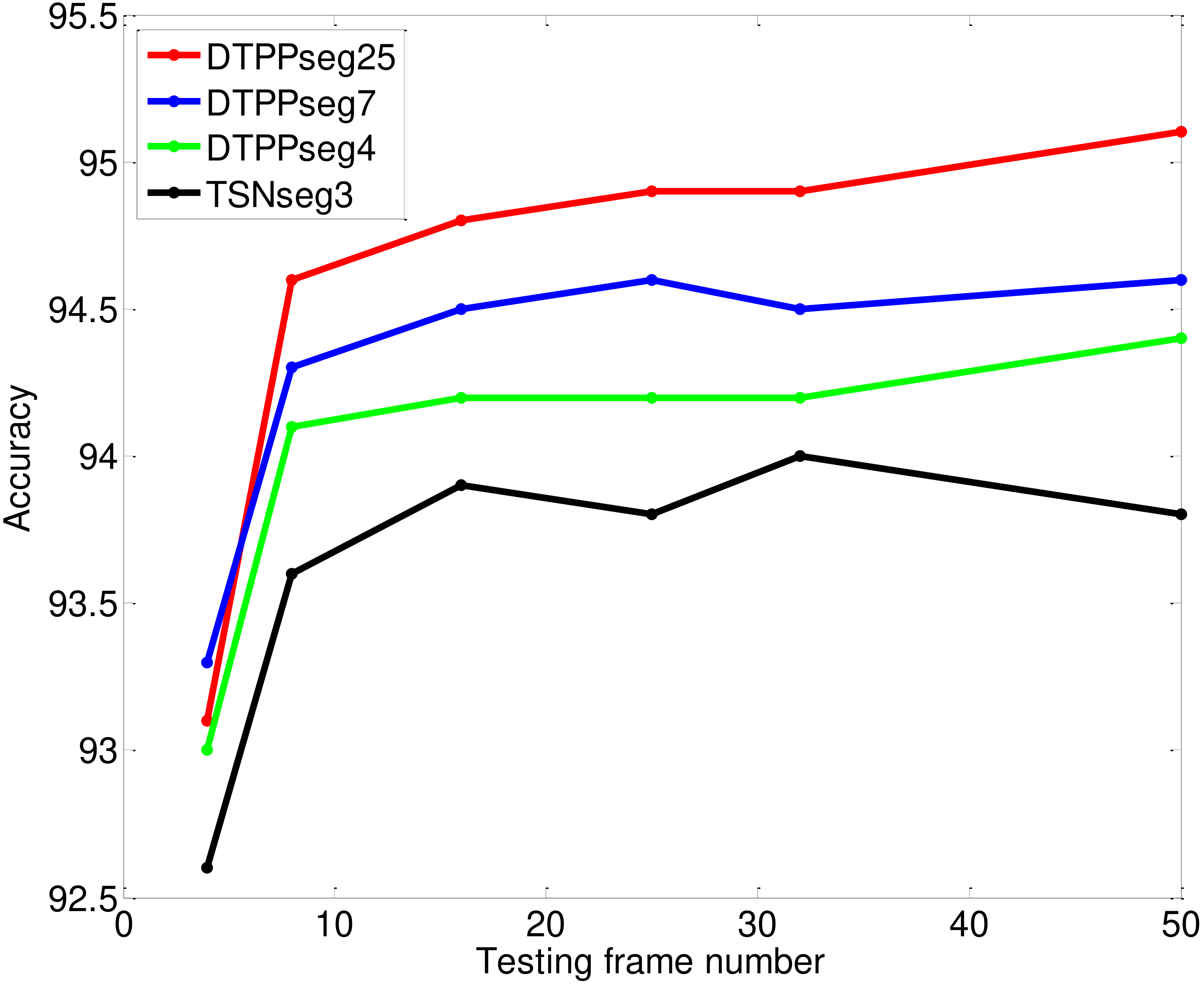}
\caption{Exploration of different number of testing frames for TSN~\cite{TSN16}, and different number of training and testing frames for DTPP on UCF101 (split 1), accuracy({\%}).}
\label{Figure_TSN_DTPP}
\end{figure}

\textbf{With different number of frames for training and testing.} Due to the nature of temporal pyramid pooling, when the temporal dimension of input video frames varies, the model could still generate a fixed-size representation. Hence in testing, it could generalize to frame number which is different from training frame number per video. This is different from some video-level representation learning methods, \eg TLE~\cite{TLE16} and DOVF~\cite{DOVF17}, which need fixed-size input in testing. While the frame-level feature learning approaches, such as two-stream ConvNets~\cite{TwoStream14} and TSN~\cite{TSN16,TSN17}, suffer from the sub-optimal mapping caused by the late fusion of the prediction scores. Our method based on temporal pyramid pooling could generalize to less or more frames during testing in representation level, instead of prediction level. In Figure~\ref{Figure_TSN_DTPP}, we show the performance of DTPP under different number of training frames per video, \eg ``DTPPseg4'' is trained by sampling 4 frames from each video. ``TSNseg3'' is the TSN~\cite{TSN16} trained by sampling 3 frames from each video. Under different number of testing frames, the accuracy of DTPP is generally stable and consistently higher than that of TSN. Specifically, ``DTPPseg4'' outperforms ``TSNseg3'' under all the testing frames cases. It clearly suggests that building the video-level representation could avoid the mistakes that the frame-level methods easily make, even using less frames for training. It could also be observed that with more frames each video for training, the performance of DTPP could be improved, showing the significance of using enough frames to build the video-level representation.


\subsection{Comparison with the state-of-the-art}

\begin{table*}
\begin{center}
\begin{tabular}{|c|c|c|c|c|c|}
\hline
               & Spatial & Temporal & Two-stream & Two-stream + MIFS & Two-stream + iDT \\
\hline\hline
Split 1        & 89.7    & 89.1     & 94.9       & 95.6              & 95.6 \\
Split 2        & 90.3    & 91.6     & 96.2       & 96.7              & 96.8 \\
Split 3        & 89.1    & 92.5     & 96.3       & 96.0              & 96.3 \\
\hline
\end{tabular}
\end{center}
\caption{The performance of DTPP on the three testing splits of UCF101, accuracy({\%}).}
\label{Table5}
\end{table*}

\begin{table*}
\begin{center}
\begin{tabular}{|c|c|c|c|c|c|c|}
\hline
               & Spatial & Temporal &\tabincell{c}{Two-stream\\(0.6 for temporal stream)} & \tabincell{c}{Two-stream\\(0.5 for temporal stream)}& \tabincell{c}{Two-stream\\+ MIFS} & \tabincell{c}{Two-stream\\+ iDT} \\
\hline\hline
Split 1        & 61.5	 & 66.3	& 75.0	         &  74.8	      & 76.9	          & 76.3 \\
Split 2        & 61.2	 & 69.2	& 75.0	         &  74.1	      & 76.3	          & 74.6 \\
Split 3        & 60.5	 & 68.8	& 74.4	         &  73.8	      & 75.9	          & 75.1 \\
\hline
\end{tabular}
\end{center}
\caption{The performance of DTPP on the three testing splits of HMDB51, accuracy({\%}).}
\label{Table6}
\end{table*}

\begin{table}
\begin{center}
\begin{tabular}{|c|c|c|}
\hline
                                                  & UCF101    & HMDB51 \\
\hline\hline
IDT~\cite{IDT13}                                  & 85.9       & 57.2           \\
MoFAP~\cite{MOFAP16}                              & 88.3       & 61.7       \\
MIFS~\cite{MIFS15}                                & 89.1       & 65.1    \\
\hline\hline
Two-stream~\cite{TwoStream14}                     & 88.0       & 59.4   \\
TDD~\cite{TDD15}                                  & 90.3       & 63.2    \\
C3D (3 nets)~\cite{C3D15}                         & 85.2       & -    \\
FstCN~\cite{FstCN15}                              & 88.1       & 59.1 \\
LTC~\cite{LTC17}                                  & 91.7       & 64.8 \\
TSN(3 seg)~\cite{TSN16}                           & 94.2	   & 70.7    \\
Gated TSN~\cite{Gating17}                                        & 94.5       & - \\
TSN(7 seg)~\cite{TSN17}                           & 94.9	   & 71.0    \\
DOVF~\cite{DOVF17}                                & 94.9	   & 71.7    \\
ActionVLAD~\cite{ActionVLAD17}                    & 92.7       & 66.9    \\
Deep Quantization~\cite{DeepQuantization17}       & 94.2	   &  -       \\
ST-ResNet~\cite{STResnet16}                       & 93.4	   & 66.4    \\
ST-Multiplier~\cite{STM17}                        & 94.2	   & 68.9    \\
ST-Pyramid Network~\cite{SPN17}                   & 94.6	   & 68.9    \\
TLE~\cite{TLE16}                                  & 95.6	   & 71.1    \\
Four-Stream~\cite{DynamicImage17}                 & 95.5       & 72.5    \\
\textbf{DTPP (ours)}                              & \textbf{95.8}	   & \textbf{74.8}    \\
\hline\hline
ST-ResNet + iDT~\cite{STResnet16}                 & 94.6	   & 70.3    \\
ST-Multiplier + iDT~\cite{STM17}                  & 94.9	   & 72.2    \\
DOVF + MIFS~\cite{DOVF17}                         & 95.3	   & 75.0    \\
ActionVLAD + iDT~\cite{ActionVLAD17}              & 93.6       & 69.8    \\
Deep Quantization + iDT~\cite{DeepQuantization17} & 95.2       &   -      \\
Eigen TSN + iDT~\cite{Eigen17}                    & 95.8       & - \\
Four-Stream + iDT~\cite{DynamicImage17}           & 96.0       & 74.9    \\
\textbf{DTPP + MIFS (ours)}                       & 96.1	   & \textbf{76.3}    \\
\textbf{DTPP + iDT (ours)}                        & \textbf{96.2}	   & 75.3    \\
\hline\hline
\multicolumn{3}{|c|}{{Kinetics pre-training}}                      \\
\hline\hline
TSN(BN-Inception)~\cite{TSN_Kinetics17} & 97.0    &   -      \\
TSN(Inception v3)~\cite{TSN_Kinetics17} & 97.3    &   -      \\
Two-Stream I3D~\cite{Kinetics17}        & \textbf{98.0}	   & 80.7    \\
\textbf{DTPP (ours)}                             & \textbf{98.0}    & \textbf{82.1}        \\
\hline
\end{tabular}
\end{center}
\caption{Comparison with the state-of-the-art, accuracy({\%}).}
\label{Table7}
\end{table}

\begin{figure*}[t]
\centering
\includegraphics[width=1\linewidth]{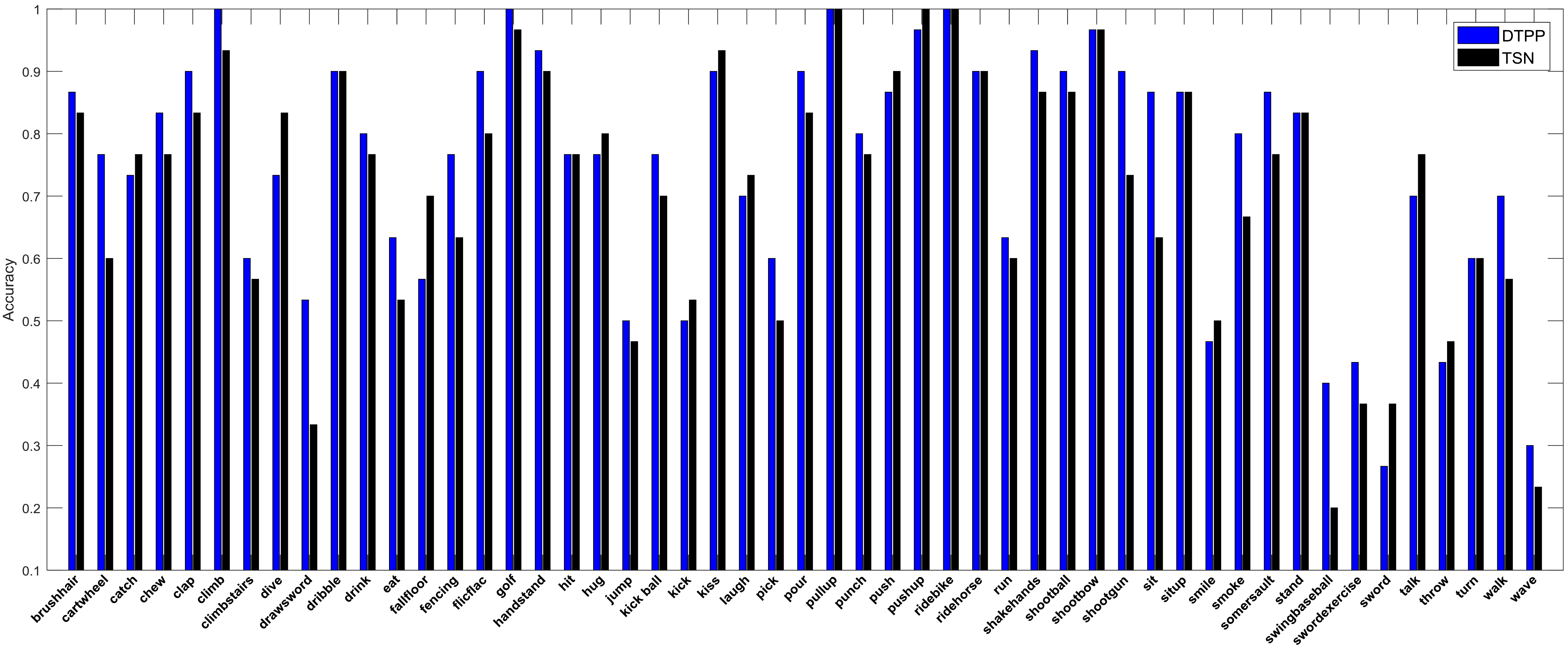}
\caption{Accuracy for each action on the testing set of HMDB51 (split 1).}
\label{Figure_hmdb}
\end{figure*}

\begin{figure*}[t]
\centering
\includegraphics[width=1\linewidth]{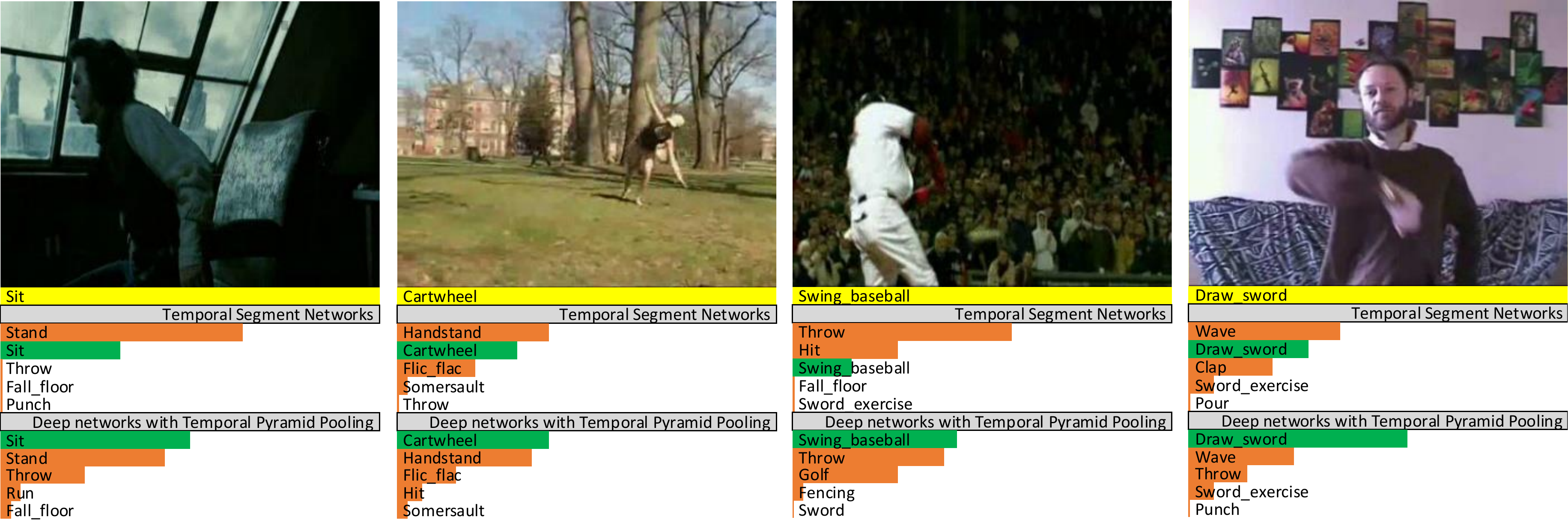}
\caption{A comparison of top-5 predictions between the Temporal Segment Networks (TSN)~\cite{TSN16} and our Deep networks with Temporal Pyramid Pooling (DTPP) on HMDB51. The yellow bars stand for the ground truth label. The green and orange bars indicate correct and incorrect predictions respectively and the length of each bar shows its confidence.}
\label{Figure_classes}
\end{figure*}

After the above analysis of DTPP, final experiments on all the three testing splits of UCF101 and HMDB51 are implemented with our proposed methods. Specifically, during training, 25 frames/short clips are sampled from each video and a temporal pooling layer of 3 levels and max pooling kernel is employed. During testing, 25 frames are selected from each video. For the late fusion of stream scores, the weights for each stream are equal, unless otherwise specified.

The performances on UCF101 and HMDB51 are reported respectively in Table~\ref{Table5} and~\ref{Table6}. As shown in Table~\ref{Table5}, the two-stream accuracy of DTPP increases from the split 1 to split 3 of UCF101, largely because of the increasing amounts of training videos, \ie 9,537, 9,586 and 9,624 respectively. The two-stream late fusion could greatly improve the accuracy over single stream. We also empirically find that with our video-level representation learning method, a equal weight (with weight 0.5) for the two-stream late fusion ensures the highest accuracy, instead of a higher weight for the temporal stream in former work, \eg 0.6 in TSN~\cite{TSN16} and 2/3 in two-stream ConvNets~\cite{TwoStream14}. By using more RGB images, our video-level representation learning could narrow the performance gap between the spatial and temporal stream. Further late fusion (with weight 0.5) with the traditional methods, MIFS~\cite{MIFS15} and iDT~\cite{IDT13} is helpful. Partly because of lacking training videos, the performance of DTPP on HMDB51 shown in Table~\ref{Table6} is less satisfied than that on UCF101. One more reason is that the videos of HMDB51 are more difficult than that of UCF101. In late fusion, a bigger weight (0.6) for the temporal stream than the spatial one gains higher accuracy than equal weights, showing the fact that the motion information of HMDB51 is more important than that of UCF101.

In Table~\ref{Table7}, the mean average accuracy on the three testing splits of UCF101 and HMDB51 of DTPP is reported. DTPP is compared with both traditional approaches \cite{MIFS15,IDT13,MOFAP16} and deep learning methods~\cite{TwoStream14,TSN16,TSN17,FstCN15,ActionVLAD17,LTC17,C3D15,TLE16,DOVF17,STResnet16,DeepQuantization17,Gating17,
TDD15,SPN17,STM17,DynamicImage17}. For single model without ensemble technique, DTPP outperforms previous best method, by 0.2\% on UCF101. Impressively, DTPP exceeds previous single model by a large margin of 2.3\% on HMDB51, which implies DTPP excels in difficult datasets where the motion cues play key role. Single model of DTPP even outperforms TSN~\cite{TSN16,TSN17} with three input modalities. With further fusion with iDT~\cite{IDT13} and MIFS~\cite{MIFS15}, DTPP achieves a new level of accuracy both on UCF101 and HMDB51, 96.2\% and 76.3\% respectively. Specifically, DTPP exceeds the frame-level feature learning method TSN~\cite{TSN16,TSN17}, the video-level representation learning method TLE~\cite{TLE16} with only 3 frames, the video feature learning method without end-to-end training and without multi-scale temporal modeling, DOVF~\cite{DOVF17}. More surprisingly, DTPP even exceeds Dynamic Image Networks~\cite{DynamicImage17} either by single model or fusion with iDT~\cite{IDT13}, which employs the ensemble of four models, all the frames of testing videos and a more advanced network architecture, \ie ResNeXt~\cite{Resnetxt16}. Dynamic Image Networks~\cite{DynamicImage17} is also a video-level representation learning method. However, it is limited in single temporal scale modeling and its encoded dynamic inputs may lose the cues of original frames~\cite{DynamicImage17}.

DTPP is also compared with other methods pre-trained on Kinetics dataset~\cite{Kinetics17}. As shown in Table~\ref{Table7}, on the UCF101 dataset, DTPP pre-trained on Kinetics exceeds TSN with BN-Inception as backbone by 1.0\% and TSN with Inception v3~\cite{Inception_v3_15} by 0.7\%. DTPP is on par with Two-stream I3D~\cite{Kinetics17} on UCF101, but exceeds it on HMDB51 by 1.4\%, thereby achieving the state-of-the-art performance when pre-trained on Kinetics.

\subsection{Visualization}

Because of the large improvements of DTPP over previous methods on HMDB51, visualizations are done on the testing set (split 1) of this dataset. Corresponding results of TSN~\cite{TSN16} are also displayed for clear illustration.

Figure~\ref{Figure_hmdb} shows the accuracy of each action. For most actions, the accuracy of our DTPP is higher than that of the TSN. For example, for ``sit'', ``cartwheel'', and ``swing baseball'', `draw sword'', the accuracy of DTPP is more than 15\% higher than that of TSN. Moreover, these classification results are shown in Figure~\ref{Figure_classes}. It is noticed that DTPP performs better than TSN in the videos with sequence pattern, \eg ``sit'' and ``stand'', ``cartwheel'' and ``handstand''. One possible reason is that the video-level representation of DTPP is sequence-aware, which is lacked in TSN. DTPP also behaves well when the human object interaction pattern exists in videos, \eg ``swing baseball'', ``draw sword''. Though we don't have specific designs on detecting detailed objects, DTPP seems to capture the fine-grained objects by using contextual information during building the video-level representation.

\section{Conclusion}
~\label{Conclusion}
In this paper, we propose Deep networks with Temporal Pyramid Pooling (DTPP), an end-to-end video-level representation learning approach. DTPP uses a temporal pyramid pooling layer to aggregate the frame-level features of videos into a multi-scale video-level representation, being both global and sequence-aware. A set of problems are studied, including the number of pyramid levels and pooling kernel, the effect of end-to-end training as well as the effect of training frame number and testing frame number per video. Finally, DTPP achieves the state-of-the-art performance on UCF101 and HMDB51, either by ImageNet pre-training or Kinetics pre-training. Though the video-level representation is obtained by end-to-end training in this work, a more thorough extension is to calculate the optical flow in a network fashion~\cite{FlowNet2.0,FlowNet1.0}, where the flownet could also be fine-tuned. A unified spatial-temporal video-level representation could also be obtained by the fusion ~\cite{convfusion16,STM17} of the feature maps from both streams.

{\small
\bibliographystyle{ieee}

}
\end{document}